\documentclass[sigconf, screen]{acmart}

\setcopyright{none}
\renewcommand\footnotetextcopyrightpermission[1]{}
\AtBeginDocument{%
  }

\setcopyright{acmlicensed}
\copyrightyear{2018}
\acmYear{2018}
\acmDOI{XXXXXXX.XXXXXXX}
\acmISBN{978-1-4503-XXXX-X/2018/06}

\acmSubmissionID{1741}

\usepackage{xcolor}
\usepackage{mdframed}
\usepackage{tcolorbox}
\tcbuselibrary{breakable} 
\tcbuselibrary{skins}     

\usepackage{enumitem}
\usepackage{colortbl}
\usepackage{booktabs}
\begin{document}

\title{Evaluating Remote Sensing Image Captions Beyond Metric Biases}

\author{Ziyun Chen}
\affiliation{
  \institution{Hohai University}
  \city{Nanjing}
  \country{China}
}
\email{hhu-czy@hhu.edu.cn}

\author{Fan Liu}
\authornote{Corresponding author.}
\affiliation{
  \institution{Hohai University}
  \city{Nanjing}
  \country{China}
}
\email{fanliu@hhu.edu.cn}

\author{Liang Yao}
\affiliation{
  \institution{Hohai University}
  \city{Nanjing}
  \country{China}
}
\email{liangyao@hhu.edu.cn}

\author{Chuanyi Zhang}
\affiliation{
  \institution{Hohai University}
  \city{Nanjing}
  \country{China}
}
\email{20231104@hhu.edu.cn}

\author{Yuye Ma}
\affiliation{
  \institution{Hohai University}
  \city{Nanjing}
  \country{China}
}
\email{2206010103@hhu.edu.cn}

\author{Wei Zhou}
\affiliation{
  \institution{Cardiff University}
  \city{Cardiff}
  \country{UK}
}
\email{zhouw26@cardiff.ac.uk}

\renewcommand{\shortauthors}{Trovato et al.}

\begin{abstract}
  The core objective of image captioning is to achieve lossless semantic compression from visual signals into textual modalities. However, the reliance on manually curated reference texts for evaluation essentially forces models to mimic specific human annotation styles, thereby masking the true descriptive capabilities of advanced foundation models. This systemic misalignment prompts a critical question: Is task-specific fine-tuning truly necessary for Remote Sensing Image Captioning, or is the perceived performance gap merely an artifact of flawed evaluation criteria? To investigate this discrepancy, we propose ReconScore, a novel reference-free evaluation metric. Rather than computing textual similarities, we assess caption quality by its capability to reconstruct the original visual elements solely from the generated text, effectively neutralizing human annotation biases. Applying this metric, we uncover a profound, counterintuitive truth: inherently powerful, unfine-tuned MLLMs surpass their fine-tuned counterparts in authentic zero-shot RSIC tasks. Driven by this structural discovery, we introduce RemoteDescriber, a completely training-free generation methodology. By employing ReconScore as a self-correction mechanism, we iteratively refine the semantic precision of MLLM outputs without any computational fine-tuning overhead. Comprehensive experiments demonstrate that RemoteDescriber achieves state-of-the-art performance on three datasets. Furthermore, we validate ReconScore's reliability and analyze the flaws of traditional metrics. Our code is available at \url{}{https://github.com/hhu-czy/RemoteDescriber}.
\end{abstract}



\maketitle
\pagestyle{plain}

\section{Introduction}
Remote Sensing Image Captioning (RSIC) serves as a fundamental bridge between raw Earth observation data and human-interpretable semantic intelligence~\cite{qu2016deep, liu2022remote, lu2017exploring, gao2025multi}. Recently, the advanced vision-language reasoning capabilities of Multimodal Large Language Models (MLLMs) have sparked a paradigm shift in this domain~\cite{chen2024makes, zhang2025sc, peng2025patch, li2026rsvg, li2024language, yao2026remoteagent}. To fully exploit this potential, prior works typically heavily fine-tune MLLMs on domain-specific datasets~\cite{zhan2025skyeyegpt, muhtar2024lhrs, bazi2024rs, zhang2024earthmarker, lin2025rs, wang2024ringmogpt}. Driven by this standard training paradigm, numerous adapted models have consistently improved the state-of-the-art (SOTA) performance on classic RSIC benchmarks.
\begin{figure}
    \centering
    \includegraphics[width=1\linewidth]{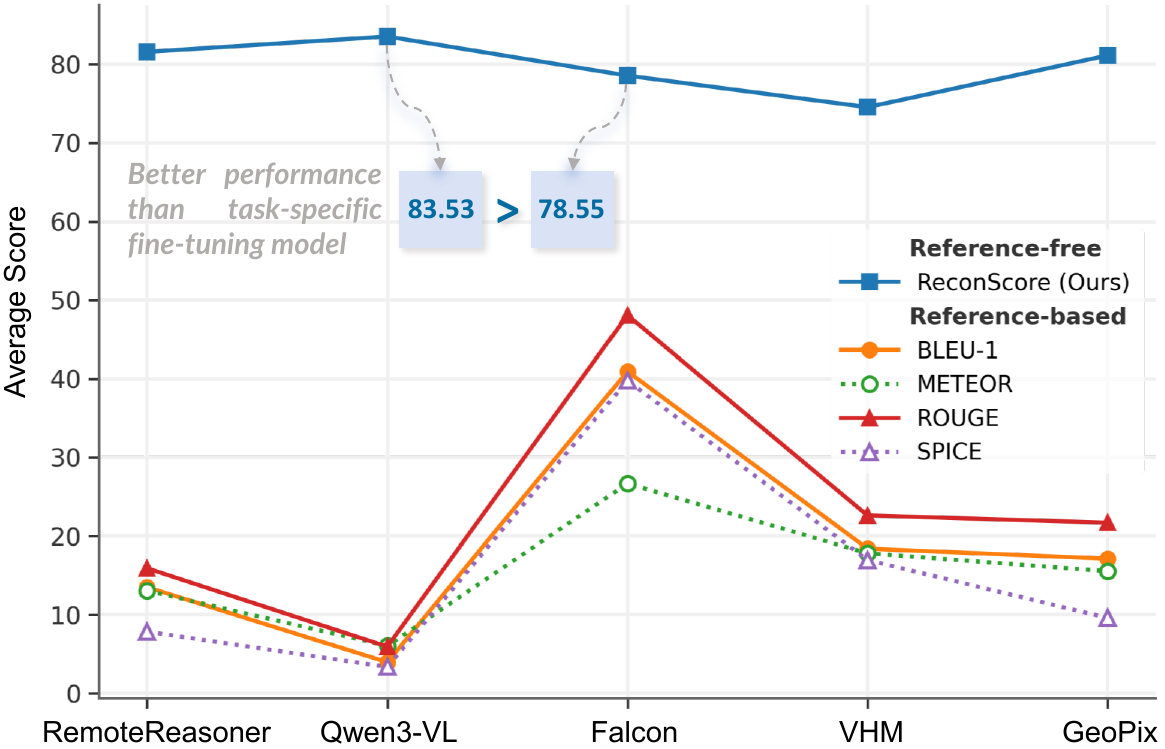}
    \caption{Comparison of different MLLMs' image captioning performance on the Sydney dataset with reference-free and reference-based metrics.}
    \label{fig:teaser}
\end{figure}

However, a key limitation of current RSIC evaluation is its reliance on human-annotated references. Given the complexity and dense object distributions in remote sensing imagery~\cite{liu2025boost, yao2024domain, yao2025uemm, zou2025remotetrimmer, jiang2026airnavigation}, these references are often sparse, capturing only a fraction of the visual content. Traditional n-gram-based metrics (e.g., BLEU~\cite{papineni2002bleu}, METEOR~\cite{banerjee2005meteor}) primarily measure textual overlap with these references. Consequently, they tend to penalize the inherently detailed and diverse MLLM-style descriptions. As illustrated in Fig.~\ref{fig:teaser} (red lines), general models (e.g., Qwen3-VL~\cite{Qwen3-VL}) score poorly on these metrics, while fine-tuned models (e.g., Falcon~\cite{yao2025falconremotesensingvisionlanguage}) achieve superficially higher scores. This evaluation paradigm encourages models to adopt shorter, simplified annotation styles during fine-tuning, which may obscure their native descriptive capabilities~\cite{urbanek2024picture}. These observations raise a crucial question: Does task-specific fine-tuning truly enhance MLLMs' visual comprehension in RSIC, or do flawed reference-based metrics simply undervalue the native descriptive capabilities of MLLMs?

Investigating this discrepancy objectively requires a reliable reference-free evaluation metric. While existing reference-free metrics, such as CLIPScore~\cite{hessel2021clipscore}, attempt to assess text-image alignment through cross-modal feature similarity, they suffer from critical limitations in the context of advanced MLLMs. Beyond strict token limits, global cross-modal embeddings often obscure fine-grained errors and missing visual details, failing to explicitly expose semantic flaws. To address these limitations, we propose that mapping the caption back into the visual domain via image reconstruction provides a more intuitive measure of the cross-modal semantic gap. Such a metric would accommodate the diverse linguistic styles of MLLMs without unwarranted penalties, offering a fair basis to re-examine the role of fine-tuning.


In this paper, we propose ReconScore, a reference-free evaluation metric driven by visual reconstruction. Rather than relying on superficial textual overlap, ReconScore assesses caption quality by measuring perceptual similarity between the original remote sensing image and the scene reconstructed from the caption. This reverse-mapping mechanism essentially highlights the factual visual elements present in the text. Additionally, this paradigm assigns appropriate credit to comprehensive descriptions without bias toward specific annotation lengths or styles. We utilize ReconScore to benchmark a wide range of MLLMs. As illustrated in Fig.~\ref{fig:teaser}, without any RSIC task-specific fine-tuning, Qwen3-VL~\cite{Qwen3-VL} exhibits highly competitive RSIC capabilities compared to its fine-tuned counterparts (e.g., Falcon~\cite{yao2025falconremotesensingvisionlanguage}). This finding suggests that the extensive visual-language alignment natively embedded in MLLMs is remarkably effective for RSIC, prompting a reconsideration of the default fine-tuning paradigm.

Driven by this discovery, we introduce RemoteDescriber, a training-free methodology for RSIC. This method employs ReconScore as a self-filtering mechanism to select the optimal, accurate descriptions from the MLLM's outputs. Due to the train-free setting, it completely avoids the high computational cost of fine-tuning, while fundamentally preserving the inherent zero-shot generalization capabilities of MLLMs. 
The experiments demonstrate the reliability of ReconScore on the preference dataset we annotated, UCM-preference. The results report that ReconScore achieves a Kendall $\tau_b$ of 28.75 and a $\tau_c$ of 35.19 over CLIPScore and other traditional reference-based metrics. In addition, the ReconScore-driven training free method RemoteDescriber outperforms the baseline MLLM, Qwen3-VL-8B, without any additional fine-tuning. In addition, we further analyze the advantages of ReconScore in semantic and length robustness.


The contributions are summarized as follows:
\begin{itemize}
    \item We propose a novel reference-free evaluation metric called ReconScore. By reconstructing images from text, it achieves objective caption evaluation that adapts to MLLMs' output. 
    \item Holistic evaluations of several MLLMs demonstrate that MLLMs can describe remote sensing images well without any additional task-specific fine-tuning for RSIC.
    \item We introduce RemoteDescriber, a training-free captioning method that can generate high-quality remote sensing image captions by leveraging zero-shot capabilities of MLLMs.

\end{itemize}
\section{Related Work}
\subsection{Remote Sensing Image Captioning}
Remote sensing image captioning (RSIC) aims to understand earth observation imagery at the semantic level and express it using natural language. Early works predominantly adopted the encoder-decoder framework. Qu et al.~\cite{qu2016deep} first introduced this framework for RSIC, combining CNN and RNN to generate remote sensing image captions. Li et al.~\cite{li2021recurrent} designed a Recurrent Attention and Semantic Gate (RASG) framework to extract effective visual information from complex geospatial scenes.

Recently, the profound vision-language reasoning capabilities of Multimodal Large Language Models (MLLMs) have propelled the development of domain-tailored models.
RSGPT~\cite{hu2025rsgpt} pioneered the introduction of MLLMs into the remote sensing field, utilizing human-annotated datasets to train a model specifically dedicated to remote sensing analysis. SkyEyeGPT~\cite{zhan2025skyeyegpt}, EarthGPT~\cite{zhang2024earthgpt}, and EarthDial\cite{soni2025earthdial} further extended the models' ability to other tasks, while also achieving good performance on the RSIC task through task-specific fine-tuning. Although these models achieved good performance on several classic benchmarks, they tend to mimic simple human annotations rather than generating comprehensive descriptions. Unlike these traditional training paradigms, GeoPixel~\cite{shabbir2025geopixel} combined pixel-level perceptual data with detailed annotation, generating rich and accurate captions. RemoteReasoner~\cite{yao2026remotereasoner}, acting as a reasoning grounding model, 
explored zero-shot captioning without specific adaptation, further demonstrating that comprehensive image captioning is an inherent capability of MLLMs.

\subsection{Image Captioning Evaluation}
Early evaluation metrics were primarily based on N-gram matching. For instance, BLEU~\cite{papineni2002bleu}, METEOR~\cite{banerjee2005meteor}, ROUGE~\cite{lin2004rouge}, and CIDEr~\cite{vedantam2015cider} evaluated captions by calculating the textual overlap between generated outputs and human-annotated references. Despite their computational efficiency, these methods focused more on lexical text similarity rather than semantic similarity. In contrast, the subsequent SPICE~\cite{anderson2016spice} metric was evaluated from the perspective of scene graphs. BERTScore~\cite{zhang2019bertscore} calculated token similarity through contextual embeddings, and CAPTURE~\cite{dong2024benchmarking} achieved caption evaluation by extracting and coupling core information. However, these metrics still rely on reference quality without directly aligning with images, while the annotations of RSIC benchmarks are typically sparse and simple, with human biases.


To circumvent the limitations and annotation costs associated with using reference for evaluation, researchers introduced reference-free evaluation metrics. UMIC~\cite{lee2021umic} was trained using contrastive learning and is capable of discriminating negative captions. CLIPScore~\cite{hessel2021clipscore} and PAC-S~\cite{sarto2023positive} evaluated image-caption similarity based on the CLIP model. Although CLIP-based metrics have been adopted in recent remote sensing works~\cite{chen2025integrating, yao2025remotesam}, their strict input token limitations render them architecturally unsuitable for evaluating long-form, dense descriptions.  Fleur~\cite{lee2024fleur} uses MLLMs to compare the caption to the image without references, which can introduce hallucinations and model biases. Human-grade evaluation was used to assess MLLM-style captions in RSGPT~\cite{hu2025rsgpt}, leading to high manual costs. These paradigms primarily focus on direct alignment between images and text, ignoring the measurement of the cross-modal information gap.


\begin{figure*}[htbp]
    \centering
    \includegraphics[width=\textwidth]{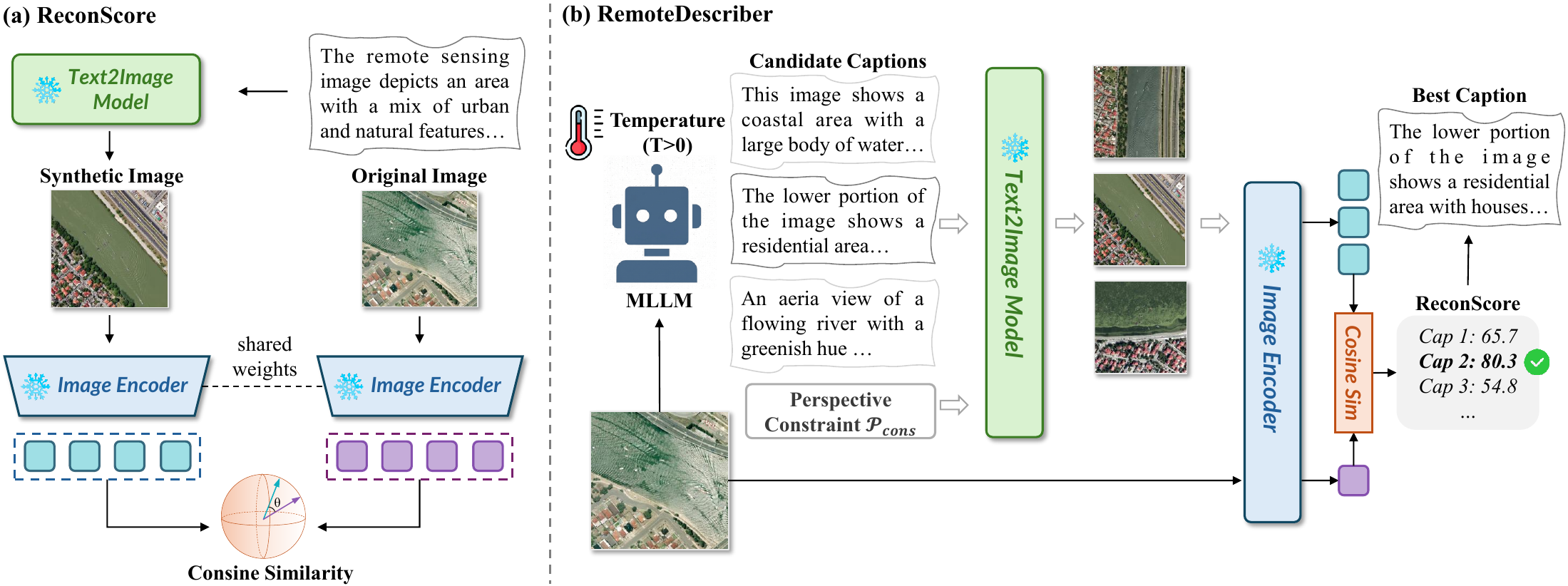}
    \caption{Overview of our method. (a) The ReconScore is computed as the cosine similarity between the reconstructed image and the original image. (b) Through ranking the candidate captions with ReconScore, RemoteDescriber can select the best caption as the final output without additional training cost. }
    \label{fig:placeholder}
\end{figure*}

\section{ReconScore}
Existing n-gram-based evaluation paradigms for RSIC typically rely on superficial textual overlap, failing to reliably evaluate rich and diverse MLLM-generated captions. To address this limitation, we introduce ReconScore, a reference-free RSIC evaluation metric through visual reconstruction, which anchors the assessment directly to the objective visual facts of the image. 
\subsection{Theoretical Motivation}
\label{sec:formulation}
We first formulate the image captioning task and its evaluation from an information-theoretic perspective~\cite{cover1999elements}. Given an image data space $\mathcal{I}$, the goal of image captioning is to generate a description $T$ that serves as a linguistic representation of an image $I \in \mathcal{I}$.

Ideally, a high-quality caption $T$ should capture the maximal semantic information from the image $I$. To quantify this shared semantic overlap, we consider the inherent information complexity of the image as the marginal entropy $H(I)$, and the remaining uncertainty about the image given the caption $T$ as the conditional entropy $H(I|T)$. Based on these, the Mutual Information $MI(I; T)$ between the image and its corresponding caption is formulated as:
\begin{equation}
    MI(I; T) = H(I) - H(I|T).
\end{equation}

Evaluating the quality of $T$ can thus be viewed as measuring this mutual information. Since $H(I)$ is a constant for any specific image, a higher $MI(I; T)$ corresponds to a lower conditional entropy $H(I|T)$. Therefore, the core challenge of evaluation lies in quantifying $H(I|T)$, which reflects the amount of missing or incorrect visual information in the caption. However, directly computing $H(I|T)$ is mathematically intractable due to the continuous, high-dimensional characteristic of visual space and the lack of exhaustive ground-truth annotations~\cite{blei2017variational}.


To obtain a tractable approximation, we propose evaluating the caption via visual reconstruction. Specifically, we introduce a parametric generative model $p_\theta(I|T)$ (e.g., a pre-trained text-to-image model) to approximate the true posterior $p(I|T)$. According to variational inference~\cite{kingma2013auto, rezende2014stochastic}, the conditional entropy $H(I|T)$ is upper-bounded by the cross-entropy objective:
\begin{equation}
    H(I|T) \leq \mathbb{E}_{p(I,T)} [-\log p_\theta(I|T)].
\end{equation} In practice, the tightness of this variational bound depends on the approximation gap between the parameterized model $p_\theta$ and the true distribution $p$. In a theoretical optimum, we assume an ideal generative model that acts as a perfect and deterministic mapping from the text to the visual space (i.e., $p_\theta \rightarrow p$). Under this ideal fixed mapping, the generative stochasticity and systemic bias vanish, allowing the cross-entropy objective to strictly converge to the true conditional entropy $H(I|T)$. Consequently, any variance in the evaluated score would be exclusively attributed to the informational deficiency of the caption $T$.

To make this objective measurable in a semantic space, we further assume that the normalized representations of the images follow a von Mises-Fisher (vMF) distribution on the unit hypersphere~\cite{banerjee2005clustering, wang2020understanding}. Under this assumption, the negative log-likelihood $-\log p_\theta(I|T)$ can be mathematically proportional to the cosine distance between the original image $I$ and the reconstructed image $\hat{I}$ generated by the T2I model, i.e., $1 - \cos(I, \hat{I})$. Consequently, in our evaluation paradigm, the cosine-based semantic distance functions directly as a computable proxy for the upper bound of $H(I|T)$. By measuring this reconstruction gap, we obtain an estimate of how much mutual information the caption $T$ preserves from the original visual content.

\subsection{Image Reconstruction}



To approximate the ideal fixed mapping derived above, we utilize a practical Text-to-Image (T2I) model to act as a controlled visual rendering engine. 
To effectively minimize the approximation gap in our variational bound, we adopt Z-Image~\cite{team2025zimage} as the advanced T2I model, which possesses powerful visual priors. 

\begin{table}[htbp] 
    \centering
    \caption{Prompt templates utilized in our framework. (a) Perspective constraint prompt for the Text-to-Image generation model. (b) Guidelines for the MLLM to generate comprehensive descriptions.}
    \label{tab:prompts}
    
    \begin{tcolorbox}[colback=blue!2, colframe=black!60, arc=2mm, auto outer arc, boxrule=0.5pt, left=8pt, right=8pt, top=8pt, bottom=8pt, title=\textbf{Prompt Templates}]
    \small
    
    \textbf{(a) Perspective Constraint Prompt ($\mathcal{P}_{cons}(T)$)} \\
    \textit{I want a remote sensing image with a realistic satellite perspective view.} \textbf{\{\textit{T}\}} \textit{Remember, I want a vertical remote sensing satellite perspective from top to bottom.}
    
    \vspace{3mm}
    \hrule height 0.5pt 
    \vspace{3mm}
    
    \textbf{(b) Captioning Task Prompt ($\mathcal{P}_{task}$)} \\
    \textit{You are a professional expert in remote sensing, specializing in image captioning. Given a remote sensing image, your goal is to generate an informative and highly accurate description.}
    \vspace{1mm}
    \\
    \textbf{Guidelines:}
    \begin{itemize}[leftmargin=8pt]
        \setlength{\itemsep}{2pt}
        \setlength{\parskip}{0pt}
        \item \textit{Extract key objects and visual details as comprehensively as possible.}
        \item \textit{Describe the attributes of objects in detail, including quantity, color, material, shape, size, as well as absolute and relative spatial positions.}
        \item \textit{Strictly avoid hallucinated content, inaccuracies, and irrelevant information. Highlight essential visual elements without describing subjective feelings or atmosphere.}
        \item \textit{Adopt a macro-to-micro structure: first describe the overall scene, followed by specific objects.}
        \item \textit{Ensure the output is coherent, logically structured, and concise.}
    \end{itemize}
    \end{tcolorbox}
\end{table} 

Formally, given an input caption $T$, we aim to generate the reconstructed remote sensing image $\hat{I}$ through the diffusion-based T2I model $\mathcal{G}$. To explicitly guide the model in synthesizing an accurate aerial view, we embed $T$ into a specific perspective constraint prompt template $\mathcal{P}_{cons}(\cdot)$, as shown in Tab.~\ref{tab:prompts}. Consequently, the explicit image reconstruction process is formulated as:

\begin{equation}
\hat{I} = \mathcal{G}(\mathcal{P}_{cons}(T), \theta, z),
\end{equation} where $\theta$ denotes the frozen weights of Z-Image, and $z$ signifies the initial latent noise. 
To isolate potential random variables in the T2I model, both the architecture weights $\theta$ and the noise distribution $z$ are held strictly constant across all evaluations. By aggressively controlling these generative factors as constants, the T2I model is mathematically constrained into a fixed, deterministic mapping. Therefore, any variance in the synthesized image $\hat{I}$ stems exclusively from the semantic variations in $T$.


\subsection{Similarity Evaluation}
To objectively measure the semantic equivalence between the original image $I$ and the reconstructed one $\hat{I}$, we map both visual signals in the same feature space for direct comparison. 
To align with human intuitive judgments, we utilize DreamSim~\cite{fu2023dreamsim}, denoted as $\mathcal{F}_{\phi}(\cdot)$, as our core feature extractor. 
Then we compute the cosine similarity of the features to represent the similarity between the two images:

\begin{equation}
\cos(I, \hat{I}) = \frac{\mathcal{F}_{\phi}(I) \cdot \mathcal{F}_{\phi}(\hat{I})}{{\|\mathcal{F}_{\phi}(I)\|}_2 {\|\mathcal{F}_{\phi}(\hat{I})\|}_2}.
\end{equation}

To ensure that our evaluation metric is positively correlated with the caption's quality, we do not utilize the cosine distance to directly represent our final score.
‌Leveraging the inverse linear relationship between cosine distance and cosine similarity, we adopt the latter to represent the final evaluation score. Finally, to normalize the score to a standard probability-like interval $[0, 1]$, the ReconScore is formally presented as:
\begin{equation}
S_{recon} = \frac{\cos(I, \hat{I}) + 1}{2}.
\end{equation}

\section{RemoteDescriber}
Driven by the objective evaluation of our proposed ReconScore, our experiments demonstrate that off-the-shelf MLLMs inherently possess robust remote sensing image captioning capabilities, even without task-specific fine-tuning. Motivated by this finding, we propose RemoteDescriber, a fully training-free captioning framework, aiming to further explore the descriptive potential of MLLMs.

\subsection{Caption Generation}
To guide the MLLM towards generating informative and useful captions, we utilize a carefully crafted prompting strategy. Specifically, the task guideline prompt $\mathcal{P}_{task}$ is aimed to compel the model to comprehensively mine the geographical objects in the scene, detailing their fine-grained attributes and spatial relationships, as illustrated in Tab~\ref{tab:prompts}. 

Given the original remote sensing image $I$,  the MLLM  $\mathcal{M}$ can be prompted to generate a caption $\hat{T}=\mathcal{M}(I, \mathcal{P}_{task})$.
However, recent studies have revealed that increasing the length of MLLM-generated text could introduce more hallucinations and inaccuracies. To mitigate these problems, we strategically leverage the inherent generative stochasticity of MLLMs. Rather than relying on a single, deterministic, and potentially flawed output, we generate diverse candidate captions by increasing the decoding temperature $\tau > 0$. Formally, this diversity-driven generation process constructs a caption set $\mathbf{T}$ for each image, formulated as: 
\begin{equation}
T_i \sim P_{\mathcal{M}}(\cdot \mid I, \mathcal{P}_{task}; \tau), \quad \forall i \in \{1, 2, \dots, N\},
\end{equation}
\begin{equation}
\mathbf{T} = \{T_i\}_{i=1}^N,
\end{equation}
where $P_{\mathcal{M}}$ is the conditional probability distribution of the MLLM, $N$ is the predefined number of candidates, and $T_i$ represents the $i$-th generated caption.
\subsection{Self-correction Selection}
Having obtained the diverse candidate pool $\mathbf{T}$, the subsequent critical phase is to filter out flawed outputs and select the best caption objectively. To achieve this, we utilize our proposed metric ReconScore as an explicit, gradient-free evaluator. Following the ReconScore evaluation process, we first project each candidate caption $T_i \in \mathbf{T}$ back into the visual domain $\hat{I}_i = \mathcal{G}(\mathcal{P}_{cons}(T_i), \theta, z)$. Then we compute the ReconScore of each pair of candidate images:
\begin{equation}
s_i = S_{recon}\left(I, \mathcal{G}(\mathcal{P}_{cons}(T_i), \theta, z)\right).
\end{equation}
\begin{table}[htbp]
    \centering
    \caption{Human preference validation of ReconScore compared with existing metrics on UCM-preference. All reference-based metrics obtain negative Kendall $\tau$.}
    \label{tab:human_preference}
    
    \renewcommand{\arraystretch}{1.15} 
    \setlength{\tabcolsep}{6pt}        

    \begin{tabular}{c|c|cc}
        \toprule
        \textbf{Metric} & \textbf{Publication} & \textbf{Kendall $\tau_b$} & \textbf{Kendall $\tau_c$} \\
        \midrule
        \rowcolor{gray!20} \multicolumn{4}{l}{\textit{Reference-based}} \\
        BLEU-1~\cite{papineni2002bleu} & ACL02 & -32.20 & -39.39 \\
        BLEU-2~\cite{papineni2002bleu} & ACL02 & -27.42 & -33.55 \\
        BLEU-3~\cite{papineni2002bleu} & ACL02 & -16.36 & -20.01 \\
        BLEU-4~\cite{papineni2002bleu} & ACL02 & -16.25 & -19.89 \\
        METEOR~\cite{banerjee2005meteor} & ACL05 & -21.70 & -26.56 \\
        ROUGE~\cite{lin2004rouge} & ACL04 & -31.27 & -38.25 \\
        CIDEr~\cite{vedantam2015cider} & CVPR15 & -43.59 & -52.84 \\
        SPICE~\cite{anderson2016spice} & ECCV16 & -12.35 & -14.90 \\
        
        \rowcolor{gray!20} \multicolumn{4}{l}{\textit{Reference-free}} \\
        CLIPScore~\cite{hessel2021clipscore} & EMNLP21 & 25.75 & 31.48 \\
        \rowcolor{blue!10} \textbf{ReconScore} & \textbf{-} & \textbf{28.75} & \textbf{35.19} \\
        \bottomrule
    \end{tabular}
\end{table} 
The final best caption $T^*$ can be selected through a discrete maximization operation on the ranked candidate set:
\begin{equation}
T^* = \arg\max_{T_i \in \mathbf{T}} s_i.
\end{equation} Through this ReconScore-driven paradigm, RemoteDescriber further explores the zero-shot descriptive potential of MLLMs.

        

\section{Experiments}

\subsection{Experimental Setup}
\textbf{Dataset.} To comprehensively evaluate the image captioning performance, we conducted experiments of image captioning on Sydney~\cite{qu2016deep}, RSIEval~\cite{hu2025rsgpt}, UCM~\cite{qu2016deep}, RSITMD~\cite{yuan2022exploring}, and CoTalk~\cite{shen2025chain}. 

To evaluate the human correlation of metrics, we constructed the UCM-preference dataset. For each image in the UCM test set, we randomly sampled three candidate captions generated by a diverse pool of six models (spanning zero-shot MLLMs and domain-specific fine-tuned models) to ensure broad quality variance. A panel of expert graduate researchers in remote sensing conducted a strict double-blind ranking of these candidates. The ranking protocol was rigorously governed by three objective criteria: (1) object completeness and factual accuracy; (2) fine-grained attribute richness; and (3) spatial relationship fidelity. Any contested rankings were resolved through consensus discussion, guaranteeing a highly reliable ground-truth preference benchmark.

\textbf{Metrics.} We reported reference-free metrics, including BLEU 1-4\cite{papineni2002bleu}, METEOR\cite{banerjee2005meteor}, ROUGE\cite{lin2004rouge}, CIDEr\cite{vedantam2015cider}, and SPICE\cite{anderson2016spice}. For reference-free metrics, we adopted CLIPScore\cite{hessel2021clipscore} and our proposed ReconScore. Following CLIPScore~\cite{hessel2021clipscore}, we flattened all human judgments to a single list and report rank correlation, Kendall $\tau_b$, and $\tau_c$, over the instances.

\begin{table}[htbp]
    \centering
    \caption{Captioning performance of different MLLMs across three benchmark datasets. Pub. represents Publication.}
    \label{tab:main_results}
    \renewcommand{\arraystretch}{1.15} 
    \setlength{\tabcolsep}{4.5pt}        
    \begin{tabular}{c|c|ccc}
        \toprule 
        \textbf{Method} & \textbf{Pub.} & \textbf{Sydney}& \textbf{RSIEval} & \textbf{UCM} \\
        \midrule 
        \rowcolor{gray!20} \multicolumn{5}{l}{\textit{Fine-tuned}} \\
        GeoChat~\cite{kuckreja2024geochat} & CVPR24 & 74.53 & 71.78 & 70.83 \\
        VHM~\cite{pang2025vhm} & AAAI25 & 77.83 & 75.75 & 76.94 \\
        SkySenseGPT~\cite{luo2024skysensegpt} & arXiv24 & 74.51 & 73.46 & 72.90 \\
        Falcon-0.7B~\cite{yao2025falconremotesensingvisionlanguage} & arXiv25 & 78.55 & 77.03 & 75.16 \\
        EarthDial~\cite{soni2025earthdial} & CVPR25 & 73.06 & 71.39 & 71.37 \\
        GeoPix~\cite{ou2025geopix} & GRSM25 & 79.60 & 77.34 & 77.19 \\
        DescribeEarth~\cite{li2025describeearth} & arXiv25 & 81.12 & 78.11 & 78.26 \\
        
        \rowcolor{gray!20} \multicolumn{5}{l}{\textit{Zero-shot}} \\
        GeoGround~\cite{zhou2024geoground} & arXiv24 & 70.76 & 68.28 & 67.37 \\
        GeoPixel~\cite{shabbir2025geopixel} & arXiv25 & 79.44 & 77.71 & 77.67 \\
        RemoteReasoner~\cite{yao2026remotereasoner} & AAAI26 & 81.57 & 79.23 & 79.55 \\
        Qwen2.5-VL-7B~\cite{Qwen2.5-VL} & arXiv25 & 82.22 & 79.93 & 80.03 \\
        Qwen3-VL-8B~\cite{Qwen3-VL} & arXiv25 & 83.53 & 81.48 & 81.04 \\
        \rowcolor{blue!10} \textbf{RemoteDescriber} & \textbf{-} & \textbf{85.67} & \textbf{83.54} & \textbf{83.72} \\
        \bottomrule 
    \end{tabular}
\end{table}

\textbf{Models.} All the MLLMs used in the comparison of the captioning ability are divided into two groups: 
(1) Fine-tuned models: GeoChat~\cite{kuckreja2024geochat}, VHM~\cite{pang2025vhm}, Falcon-0.7B~\cite{yao2025falconremotesensingvisionlanguage},  EarthDial~\cite{soni2025earthdial}, DescribeEarth~\cite{li2025describeearth}, et al. (2) Zero-shot models: Qwen3-VL-8B~\cite{Qwen3-VL}, GeoPixel~\cite{shabbir2025geopixel}, RemoteReasoner~\cite{yao2026remotereasoner}, et al.
We adopted the Qwen3-VL-8B instruction version~\cite{Qwen3-VL} as the image captioning model of RemoteDescriber. Z-Image~\cite{team2025zimage} is utilized as the default T2I, and DreamSim is the default feature extractor.
We utilized the Qwen3-Max API to process all pure-text tasks.
Image foundation models used in ablation studies include CLIP~\cite{radford2021learning}, RemoteCLIP~\cite{liu2024remoteclip}, SigLIP2~\cite{tschannen2025siglip}, FG-CLIP2~\cite{xie2025fg}), DinoV2~\cite{oquab2023dinov2}, and DinoV3~\cite{simeoni2025dinov3}.

\subsection{Implementation Details}
The decoding temperature of RemoteDescriber was set to 0.8, and the number of candidate captions was fixed to $N = 10$. 
We employed fixed random seeds across all T2I generation processes. Unless otherwise specified, we uniformly set the maximum input text length to 512 tokens and the denoising steps to 28 for all T2I models. 
All synthetic images strictly preserved the aspect ratio of the original images, with the maximum spatial dimension capped at 1024 pixels.  All of our experiments could be conducted on a single NVIDIA RTX 4090D GPU with 24GB of VRAM.

\subsection{Main Results}
\textbf{Human Preference Consistency.}
To validate whether the evaluation metrics truly align with human judgment, we evaluated both reference-based and reference-free image captioning metrics on our preference dataset, UCM-preference. As illustrated in Tab.~\ref{tab:human_preference}, all traditional reference-based metrics exhibit a strong negative correlation with human preferences to MLLM-style captions. 
Since MLLMs output rich descriptions that differ from the sparse ground-truths, traditional metrics assign them low scores, even though human evaluators actually prefer these comprehensive and accurate outputs. In contrast, reference-free metrics correlate positively with human judgments, indicating their suitability for evaluating MLLM-generated captions. Our proposed ReconScore achieves the highest correlation scores (28.75 for $\tau_b$ and 35.19 for $\tau_c$), surpassing CLIPScore by 11.72\%. This result confirms that evaluating MLLM-style captions through ReconScore is highly reliable, better than relying on rigid text overlap or cross-modal global feature similarity. 


\begin{table}[t]
    \centering
    \caption{Impact of paraphrased captions versus perturbed Ground Truth (GT) on different evaluation metrics. $\pm\Delta$ denotes the score difference (Paraphrased - Perturbed GT). Both kinds of captions are rewritten by prompting an LLM.}
    \label{tab:synonym_robustness}
    \renewcommand{\arraystretch}{1.15} 
    \setlength{\tabcolsep}{6pt}        

    \begin{tabular}{c|ccc}
        \toprule
        \textbf{Metric} & \textbf{Paraphrased} & \textbf{Perturbed GT} & \textbf{$\pm\Delta$} \\
        \midrule
        \rowcolor{gray!20} \multicolumn{4}{l}{\textit{Reference-based}} \\
        BLEU-1~\cite{papineni2002bleu} & 41.37 & 91.35 & \textcolor{red}{-49.98} \\
        BLEU-2~\cite{papineni2002bleu} & 24.89 & 89.56 & \textcolor{red}{-64.67} \\
        BLEU-3~\cite{papineni2002bleu} & 14.70 & 87.84 & \textcolor{red}{-73.14} \\
        BLEU-4~\cite{papineni2002bleu} & 8.65 & 86.12 & \textcolor{red}{-77.47} \\
        METEOR~\cite{banerjee2005meteor} & 21.17 & 54.30 & \textcolor{red}{-33.13} \\
        ROUGE~\cite{lin2004rouge} & 34.71 & 91.61 & \textcolor{red}{-56.90} \\
        CIDEr~\cite{vedantam2015cider} & 11.43 & 437.64 & \textcolor{red}{-426.21} \\
        SPICE~\cite{anderson2016spice} & 21.08 & 82.68 & \textcolor{red}{-61.60} \\
        
        \rowcolor{gray!20} \multicolumn{4}{l}{\textit{Reference-free}} \\
        CLIPScore~\cite{hessel2021clipscore} & 73.95 & 74.85 & \textcolor{red}{-0.90} \\
        \rowcolor{blue!10} \textbf{ReconScore} & 80.21 & 77.48 & \textbf{\textcolor{green!60!black}{+2.73}} \\

        \bottomrule
    \end{tabular}
\end{table}

\textbf{Is Fine-tuning Necessary for RSIC?}
We utilized our proposed ReconScore to objectively evaluate the captioning performance of various open-source MLLMs.
An interesting observation as shown in Tab.~\ref{tab:main_results}, the representative fine-tuned models like GeoChat and Falcon-0.7B demonstrate solid performance, but they do not show a decisive advantage under the fact-based evaluation of ReconScore. In contrast, the general pre-trained MLLM, Qwen3-VL-8B, without any remote sensing-specific fine-tuning, achieves remarkable scores of 83.53, 81.48, and 81.04 across the three datasets, exceeding many fine-tuned models. Additionally, RemoteReasoner, a reasoning visual grounding task-specific model, also yields strong results
, which may mean training on perceptual reasoning tasks also helps improve the model's vision-language understanding. 
These experimental findings offer a new perspective on the RSIC task: while task-specific fine-tuning is undeniably valuable for adapting models to the human-style annotations, it may not be strictly necessary for achieving comprehensive visual comprehension. 

\begin{table}[htbp]
    \centering
    
    \caption{Robustness analysis across caption lengths on RSITMD. \textbf{S.}, \textbf{M.}, and \textbf{L.} represent Short, Medium, and Long text categories, with average length of 18, 54.78, and 167.82 words respectively. $\sigma$ denotes standard deviation.}
    \label{tab:length_robustness_std}
    \renewcommand{\arraystretch}{1.15}
    \begin{tabular}{c|c|ccc|c}
        \toprule
        \textbf{Metric} & \textbf{Max Tokens} & \textbf{S.} & \textbf{M.} & \textbf{L.} & \textbf{$\sigma \downarrow$} \\
        \midrule
        
        \rowcolor{gray!20} \multicolumn{6}{l}{\textit{Reference-based}} \\
        BLEU-1~\cite{papineni2002bleu} & - & 75.16 & 27.74 & 9.37 & 33.95 \\
        BLEU-2~\cite{papineni2002bleu} & - & 57.67 & 18.28 & 5.53 & 27.18 \\
        BLEU-3~\cite{papineni2002bleu} & - & 41.63 & 10.46 & 2.48 & 20.69 \\
        BLEU-4~\cite{papineni2002bleu} & - & 26.81 & 4.50 & 0.62 & 14.13 \\
        METEOR~\cite{banerjee2005meteor} & - & 39.55 & 25.09 & 12.05 & 13.76 \\
        ROUGE~\cite{lin2004rouge}  & - & 56.09 & 26.05 & 10.39 & 23.22 \\
        CIDEr~\cite{vedantam2015cider}  & - & 0.00  & 0.00  & 0.00  & - \\
        SPICE~\cite{anderson2016spice}  & - & 60.12 & 30.82 & 11.83 & 24.33 \\
        
        \rowcolor{gray!20} \multicolumn{6}{l}{\textit{Reference-free}} \\
        CLIPScore~\cite{hessel2021clipscore} & 77 & 70.79 & 71.21 & 70.63 & 0.30 \\
        \rowcolor{blue!10} \textbf{ReconScore} & 512 & 75.66 & 75.78 & 74.97 & 0.44 \\
        \bottomrule
    \end{tabular}
\end{table}

\textbf{Semantic Fidelity over Lexical Overlap.}
To investigate whether evaluation metrics can distinguish genuine semantic errors from mere lexical variations, we designed a controlled perturbation experiment on the CoTalk dataset. For each Ground Truth (GT) caption, we constructed two specific variants: (1) a \textit{Paraphrased} caption, which alters the vocabulary and syntax but strictly preserves the original visual semantics; and (2) a \textit{Perturbed} caption, which retains high lexical overlap with the GT but introduces a fine-grained semantic hallucination (e.g., altering an object class or spatial relation). Logically, a robust metric must assign a higher score to the semantically faithful paraphrased caption than to the factually flawed perturbed one. As presented in Tab.~\ref{tab:synonym_robustness}, traditional reference-based metrics consistently yield massive negative $\Delta$ values, since they blindly reward the superficial word overlap of the perturbed captions.
Furthermore, the reference-free CLIPScore also fails to overcome this bias of -0.90 $\Delta$, as its global feature pooling mechanism is insensitive to localized, word-level hallucinations. In contrast, ReconScore is the only metric that achieves a positive margin ($\Delta = +2.73$). By translating textual flaws into obvious visual discrepancies, our metric demonstrates superior sensitivity to fine-grained 
semantic fidelity aligning with the image.

\subsection{Further Analysis}

\begin{figure*}[tp]
    \centering
    \includegraphics[width=\textwidth]{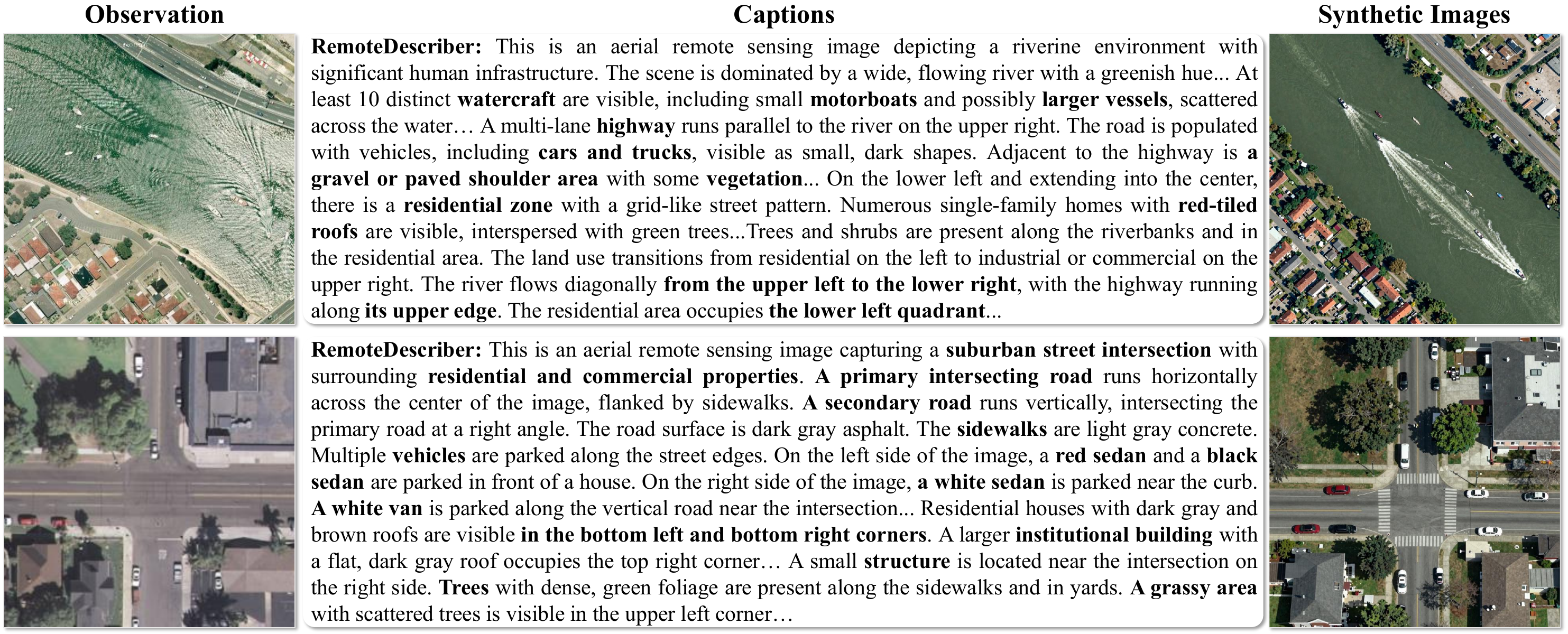}
    \caption{Visualization results of RemoteDescriber. The bold words represent the key described visual elements in the image.
    }
    \label{fig:visualization}
\end{figure*}

\textbf{Length Robustness of Different Metrics.}
 We designed a controlled generation pipeline utilizing the RSITMD dataset to evaluate the robustness of different metrics against textual length. We employed an LLM to randomly extract 10 semantic triplets (e.g., <cars, are parked at, parking lots>) from the original GT annotations. The LLM is prompted to synthesize captions of three different length categories conditioned strictly on these fixed triplets, which guarantees the same semantics. 
As shown in Tab.~\ref{tab:length_robustness_std}, reference-based scores plummet for longer captions despite preserved semantics, heavily penalized by the sparse ground-truth annotations. Conversely, reference-free metrics exhibit high stability. While CLIPScore yields the lowest standard deviation (0.30), its restrictive token limit renders it inherently unreliable for dense descriptions.
In contrast, our proposed ReconScore can accommodate much longer input tokens and maintain excellent robustness of caption length, achieving a standard deviation of 0.44. 
These experiments demonstrate that ReconScore can provide fair evaluations for MLLM-style captions without the biases of their length.

\textbf{Human Preference of Zero-Shot vs. Fine-Tuned Models.}
To empirically validate our counterintuitive finding that zero-shot MLLMs surpass task-specific fine-tuned models, we conducted a statistical analysis on the UCM-preference dataset. Based on the previously established blind-test ground truth, we calculated the overall human preference proportion for each model paradigm to reflect real visual expectations. 
The experimental results, summarized in Table \ref{tab:human_preference_score}, zero-shot MLLMs not only dominate human evaluations with an $84.10\%$ (291 out of 346) preference proportion, but also consistently receive higher average ReconScores of 80.14. This dual confirmation fundamentally demonstrates that zero-shot MLLMs inherently generate descriptions more aligned with human visual cognition for remote sensing. Furthermore, the strong consistency between our metric and human preference validates ReconScore as a reliable and objective evaluator for RSIC.

\textbf{Overfitting Risk of Metric-Driven Selection Policies.}
We conducted a human preference experiment on the selection policy of RemoteDescriber to strictly rule out the risk of metric overfitting. Different evaluation metrics are employed to independently score the candidate pool and select their respective top-ranked captions. Subsequently, expert annotators performed a double-blind assessment to choose the absolute ``best'' description from these metric-selected outputs. We measured the percentage of metric-selected captions that match the human preference.
As reported in Tab.~\ref{tab:selection_ablation}, captions selected via ReconScore achieve the highest human preference rate (50.00\%), outperforming those selected by CLIPScore (44.83\%) and METEOR (29.31\%). 
By utilizing it as an active zero-shot reward signal, RemoteDescriber steers the generated descriptions toward real human visual expectations, mitigating the concerns of reward hacking.

\begin{table}[tp]
    \centering
    \caption{Human preference evaluation of zero-shot vs. fine-tuned models. ``Wins'' indicates the frequency a model's output was selected as superior by human annotators.}
    \label{tab:human_preference_score}
     \setlength{\tabcolsep}{2.6mm}
    \renewcommand{\arraystretch}{1.1}
    \begin{tabular}{c|ccc}
    \toprule
    \textbf{Model Paradigm} & \textbf{ReconScore} & \textbf{Wins} & \textbf{Proportion} \\
    \midrule
    Fine-Tuned & 74.87 & 55 & 15.90\% \\
    \rowcolor{blue!10} \textbf{Zero-Shot} & \textbf{80.14} & \textbf{291} & \textbf{84.10\%} \\
    \bottomrule
    \end{tabular}
\end{table}

\textbf{Qualitative Analysis.} 
The qualitative results of RemoteDescriber are presented in Fig.~\ref{fig:visualization}. RemoteDescirber comprehensively describes both fine-grained visual elements and complex spatial layouts. 
This dense extraction of geospatial information ensures that the corresponding synthetic images faithfully reconstruct the original observations. 
The high degree of visual alignment between the input and reconstructed scenes demonstrates RemoteDescriber's ability to generate highly accurate, comprehensive, and precise 
spatial structures without any task-specific fine-tuning. 

\begin{table}[tp]
    \centering
    \caption{Human preference evaluation of different selection metrics on Sydney. The preference rate represents the proportion of human agreement with the best caption selected by a specific metric.}
    \label{tab:selection_ablation}
    
    \renewcommand{\arraystretch}{1.15}
    \begin{tabular}{c|ccc}
        \toprule
        \textbf{Selection Metric} & METEOR & CLIPScore & \textbf{ReconScore} \\
        \midrule
        \textbf{Preference Rate}  & 29.31\% & 44.83\% & \cellcolor{blue!10}\textbf{50.00\%} \\
        \bottomrule
    \end{tabular}
\end{table}

\subsection{Ablation Studies}

\textbf{Different Numbers of Candidate Captions.}
We conducted an ablation study investigating the influence of the candidate pool size ($N$) on the final captioning quality to evaluate the core selection mechanism of our RemoteDescriber framework. Table~\ref{tab:candidate_numbers} summarizes the performance across the Sydney, RSIEval, and UCM datasets as $N$ increases from 1 to 10. 
When $N=1$, the framework relies on a single generation pass without any selection mechanism, yielding baseline ReconScores of 83.32, 81.24, and 81.23, respectively.
As we gradually enlarged the candidate pool, the ReconScores exhibited a consistent and monotonic performance improvement. 
Notably, increasing $N$ from 1 to 10 brings substantial absolute gains of +2.35, +2.30, and +2.49 points on the three datasets, respectively.
The experimental results demonstrate that RemoteDescriber dramatically increases the probability of generating a flawless description by exploring a larger and more diverse candidate space ($N>1$).

\begin{table}[tp]
    \centering
    \caption{Influence of the number of candidate captions ($N$) utilized in RemoteDescriber.}
    \label{tab:candidate_numbers}
    
    \renewcommand{\arraystretch}{1.15} 
    \setlength{\tabcolsep}{6pt}        

    \begin{tabular}{c|ccc|c}
        \toprule
        \textbf{Candidates ($N$)} & \textbf{Sydney} & \textbf{RSIEval} & \textbf{UCM} & \textbf{Total} \\
        \midrule
        1 & 83.32 & 81.24 & 81.23 & 245.79 \\
        2 & 84.03 & 82.27 & 82.12 & 248.42 \\
        4 & 85.08 & 82.92 & 82.94 & 250.94 \\
        6 & 85.29 & 83.26 & 83.24 & 251.79 \\
        8 & 85.50 & 83.33 & 83.48 & 252.31 \\
        \rowcolor{blue!10} 10 & \textbf{85.67} & \textbf{83.54} & \textbf{83.72} & \textbf{252.93} \\
        \bottomrule
    \end{tabular}
\end{table}

\begin{figure}
    \centering
    \includegraphics[width=1\linewidth]{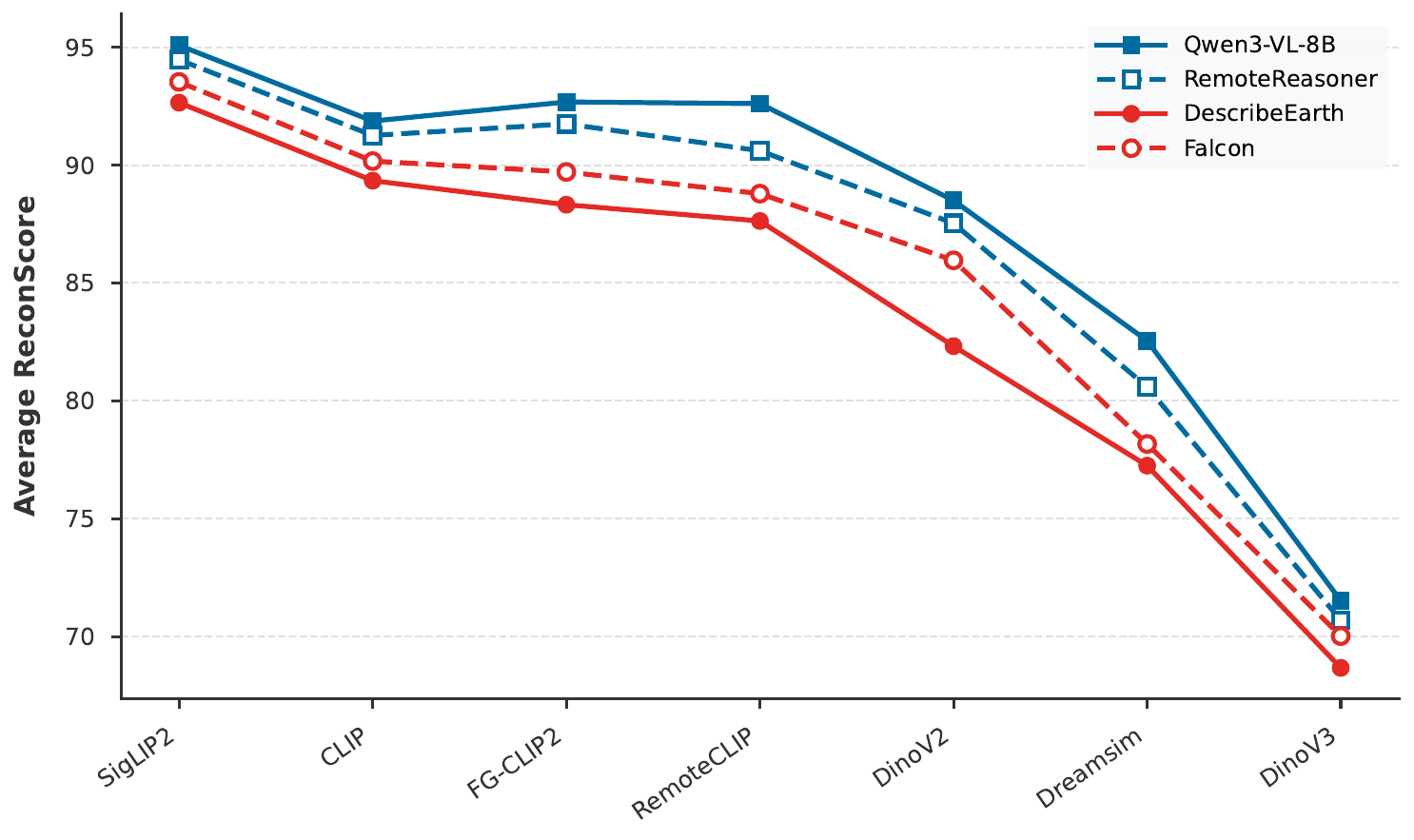}
    \caption{Comparison of different Image Encoders for RemoteDescriber on CoTalk dataset. All the ReconScores are computed by the normalized cosine similarity.}
    \label{fig:encoder}
\end{figure}


\textbf{Different Image Encoders.}
To further exclude the concern that ReconScore might merely reflect the specific inductive biases of the T2I model or the feature extractor, rather than the real semantic fidelity of the captions, we conducted an ablation on the visual evaluator. Different encoders are optimized with fundamentally different training objectives, leading to entirely distinct feature spaces and numerical bounds for cosine similarity. Consequently, the absolute score variations observed in Fig.~\ref{fig:encoder} simply reflect these baseline distributions. Notably, despite the stark differences in these perceptual spaces, the relative ranking of the evaluated MLLMs remains strictly consistent across all conditions
(where Qwen3-VL stably leads, followed by RemoteReasoner, Falcon, and DescribeEarth)
. This cross-encoder consistency validates that the semantic differences of the input captions, independent of other latent factors, fundamentally drive the variance in evaluation scores.

\begin{figure}
    \centering
    \includegraphics[width=1\linewidth]{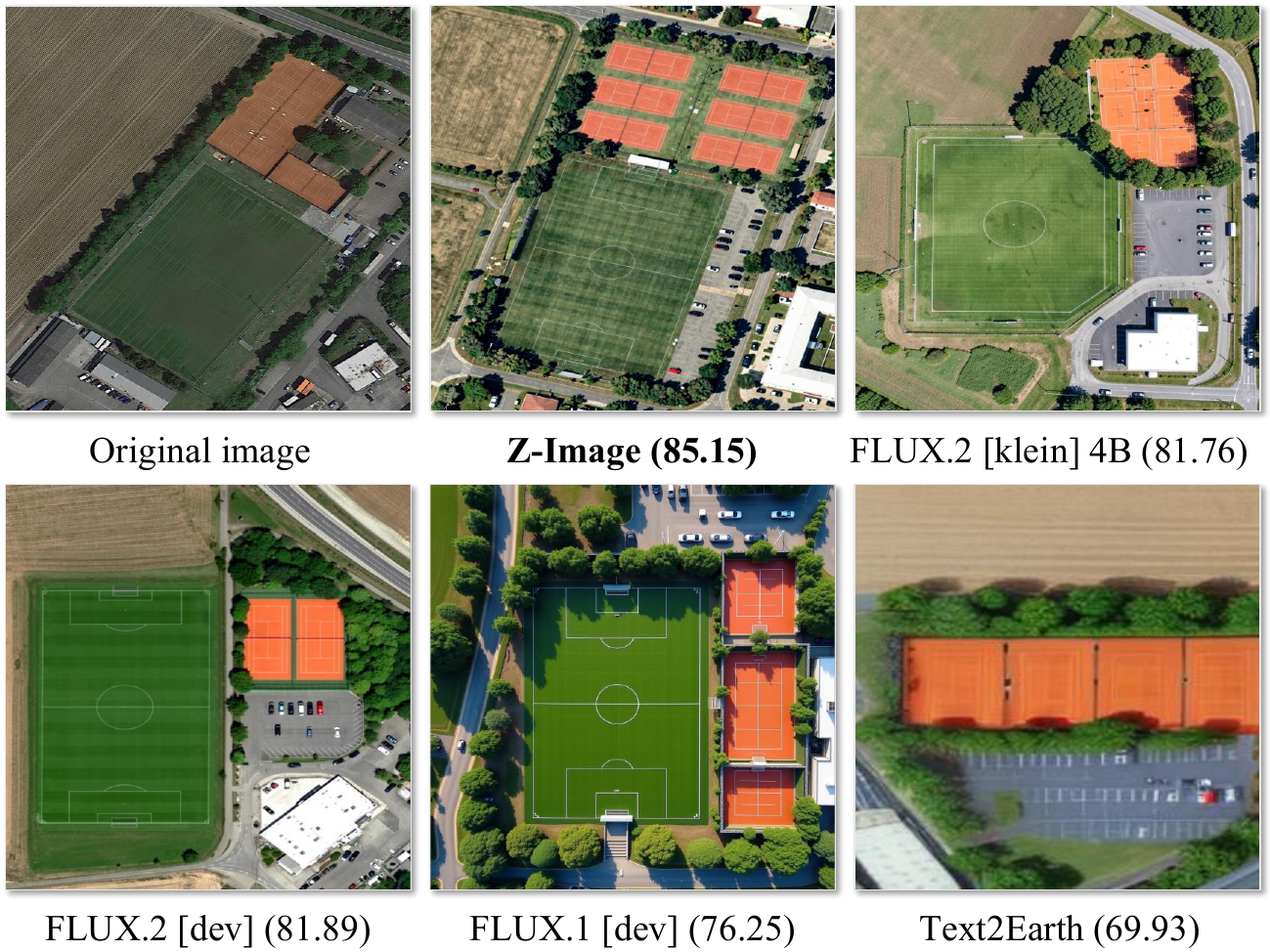}
    \caption{Synthetic images reconstructed by different models based on the same caption and ReconScores in parentheses.
    }
    \label{fig:t2i}
\end{figure}

\textbf{Different Text-to-Image Models. }
The performance of the Text-to-Image (T2I) model fundamentally dictates the reliability of ReconScore and RemoteDescriber. 
To minimize systemic bias, we ablated several T2I models to determine the most effective T2I engine for our framework. As shown in Tab.~\ref{tab:t2i_ablation}, Text2Earth yields the lowest final ReconScore by 71.71. Despite its extremely fast generation speed of 2.97s, its strict 77-token limit forces the aggressive truncation of the dense caption generated by MLLMs. In contrast, general domain T2I models have a longer context length and the capability of generating higher-quality synthetic images. Z-Image achieves the highest final captioning performance (80.38), surpassing FLUX.2 [klein] 4B by 1.99 with an acceptable loss on inference time. 
As shown in Fig.~\ref{fig:t2i}, given the same input caption, Z-Image generates remote sensing images that are more realistic and strictly text-faithful.
Consequently, we chose Z-Image as the high-fidelity image generator of RemoteDescriber. 
\begin{table}[tp]
    \centering
    \caption{Ablation study on the choice of the T2I model. The evaluation compares maximum input context Tokens (M. T.), inference time per image (I. T.), and the final ReconScore.}
    \label{tab:t2i_ablation}
    \setlength{\tabcolsep}{7.5pt}        

    \renewcommand{\arraystretch}{1.15}
    \begin{tabular}{c|cc|c}
        \toprule
        \textbf{Model} & \textbf{M. T.} & \textbf{I. T. (s)} & \textbf{ReconScore} \\
        \midrule
        Text2Earth~\cite{liu2025text2earth} & 77 & 2.97 & 71.71 \\
        FLUX.1 [dev]~\cite{flux2024} & 512 & 38.27 & 74.94 \\
        FLUX.2 [dev]~\cite{flux-2-2025} & 512 & 95.36 & 78.06 \\
        FLUX.2 [klein] 4B~\cite{flux-2-2025} & 512 & 17.96 & 78.39 \\
        \rowcolor{blue!10} 
        \textbf{Z-Image~\cite{team2025zimage}} & 512 & 27.88 & \textbf{80.38} \\
        \bottomrule
    \end{tabular}
\end{table}

\section{Limitations}
While ReconScore currently utilizes explicit image generation, the latent space reconstruction also represents a promising alternative. Furthermore, introducing the training-free RemoteDescriber is not to negate the necessity of task-specific fine-tuning, as zero-shot paradigms cannot yet fulfill the stringent precision demanded by expert-level remote sensing interpretation. The future efforts for advancing practical Earth observation tasks rely on constructing high-quality, domain-specific vision-language datasets coupled with highly reliable evaluation metrics.

\section{Conclusion} 
In this work, we propose ReconScore, a reconstruction-driven, reference-free metric that reveals the robust native RSIC capabilities of zero-shot MLLMs, rendering extensive fine-tuning unnecessary. Building on this paradigm-shifting insight, we introduce RemoteDescriber, a completely training-free framework that achieves state-of-the-art performance across three benchmarks while tightly aligning with human preferences. Our work demonstrates that pairing reference-free evaluation with zero-shot generation offers a highly efficient pathway to bridge human-interpretable semantics and complex Earth observation data.

\bibliographystyle{ACM-Reference-Format}
\bibliography{sample-base}

@inproceedings{qu2016deep,
  title={Deep semantic understanding of high resolution remote sensing image},
  author={Qu, Bo and Li, Xuelong and Tao, Dacheng and Lu, Xiaoqiang},
  booktitle={2016 International conference on computer, information and telecommunication systems (Cits)},
  pages={1--5},
  year={2016},
  organization={IEEE}
}

@article{li2021recurrent,
  title={Recurrent attention and semantic gate for remote sensing image captioning},
  author={Li, Yunpeng and Zhang, Xiangrong and Gu, Jing and Li, Chen and Wang, Xin and Tang, Xu and Jiao, Licheng},
  journal={IEEE Transactions on Geoscience and Remote Sensing},
  volume={60},
  pages={1--16},
  year={2021},
  publisher={IEEE}
}

@article{hu2025rsgpt,
  title={Rsgpt: A remote sensing vision language model and benchmark},
  author={Hu, Yuan and Yuan, Jianlong and Wen, Congcong and Lu, Xiaonan and Liu, Yu and Li, Xiang},
  journal={ISPRS Journal of Photogrammetry and Remote Sensing},
  volume={224},
  pages={272--286},
  year={2025},
  publisher={Elsevier}
}

@article{zhan2025skyeyegpt,
  title={Skyeyegpt: Unifying remote sensing vision-language tasks via instruction tuning with large language model},
  author={Zhan, Yang and Xiong, Zhitong and Yuan, Yuan},
  journal={ISPRS Journal of Photogrammetry and Remote Sensing},
  volume={221},
  pages={64--77},
  year={2025},
  publisher={Elsevier}
}

@article{zhang2024earthgpt,
  title={EarthGPT: A universal multimodal large language model for multisensor image comprehension in remote sensing domain},
  author={Zhang, Wei and Cai, Miaoxin and Zhang, Tong and Zhuang, Yin and Mao, Xuerui},
  journal={IEEE Transactions on Geoscience and Remote Sensing},
  volume={62},
  pages={1--20},
  year={2024},
  publisher={IEEE}
}

@inproceedings{yao2026remotereasoner,
  title={Remotereasoner: Towards unifying geospatial reasoning workflow},
  author={Yao, Liang and Liu, Fan and Lu, Hongbo and Zhang, Chuanyi and Min, Rui and Xu, Shengxiang and Di, Shimin and Peng, Pai},
  booktitle={Proceedings of the AAAI Conference on Artificial Intelligence},
  volume={40},
  number={14},
  pages={11883--11891},
  year={2026}
}

@article{shabbir2025geopixel,
  title={Geopixel: Pixel grounding large multimodal model in remote sensing},
  author={Shabbir, Akashah and Zumri, Mohammed and Bennamoun, Mohammed and Khan, Fahad S and Khan, Salman},
  journal={arXiv preprint arXiv:2501.13925},
  year={2025}
}

@inproceedings{papineni2002bleu,
  title={Bleu: a method for automatic evaluation of machine translation},
  author={Papineni, Kishore and Roukos, Salim and Ward, Todd and Zhu, Wei-Jing},
  booktitle={Proceedings of the 40th annual meeting of the Association for Computational Linguistics},
  pages={311--318},
  year={2002}
}

@inproceedings{banerjee2005meteor,
  title={METEOR: An automatic metric for MT evaluation with improved correlation with human judgments},
  author={Banerjee, Satanjeev and Lavie, Alon},
  booktitle={Proceedings of the acl workshop on intrinsic and extrinsic evaluation measures for machine translation and/or summarization},
  pages={65--72},
  year={2005}
}

@inproceedings{lin2004rouge,
  title={Rouge: A package for automatic evaluation of summaries},
  author={Lin, Chin-Yew},
  booktitle={Text summarization branches out},
  pages={74--81},
  year={2004}
}

@inproceedings{vedantam2015cider,
  title={Cider: Consensus-based image description evaluation},
  author={Vedantam, Ramakrishna and Lawrence Zitnick, C and Parikh, Devi},
  booktitle={Proceedings of the IEEE conference on computer vision and pattern recognition},
  pages={4566--4575},
  year={2015}
}

@inproceedings{anderson2016spice,
  title={Spice: Semantic propositional image caption evaluation},
  author={Anderson, Peter and Fernando, Basura and Johnson, Mark and Gould, Stephen},
  booktitle={European conference on computer vision},
  pages={382--398},
  year={2016},
  organization={Springer}
}

@article{zhang2019bertscore,
  title={Bertscore: Evaluating text generation with bert},
  author={Zhang, Tianyi and Kishore, Varsha and Wu, Felix and Weinberger, Kilian Q and Artzi, Yoav},
  journal={arXiv preprint arXiv:1904.09675},
  year={2019}
}

@article{dong2024benchmarking,
  title={Benchmarking and improving detail image caption},
  author={Dong, Hongyuan and Li, Jiawen and Wu, Bohong and Wang, Jiacong and Zhang, Yuan and Guo, Haoyuan},
  journal={arXiv preprint arXiv:2405.19092},
  year={2024}
}

@inproceedings{lee2021umic,
  title={UMIC: An unreferenced metric for image captioning via contrastive learning},
  author={Lee, Hwanhee and Yoon, Seunghyun and Dernoncourt, Franck and Bui, Trung and Jung, Kyomin},
  booktitle={Proceedings of the 59th Annual Meeting of the Association for Computational Linguistics and the 11th International Joint Conference on Natural Language Processing (Volume 2: Short Papers)},
  pages={220--226},
  year={2021}
}

@inproceedings{hessel2021clipscore,
  title={Clipscore: A reference-free evaluation metric for image captioning},
  author={Hessel, Jack and Holtzman, Ari and Forbes, Maxwell and Le Bras, Ronan and Choi, Yejin},
  booktitle={Proceedings of the 2021 conference on empirical methods in natural language processing},
  pages={7514--7528},
  year={2021}
}

@inproceedings{sarto2023positive,
  title={Positive-augmented contrastive learning for image and video captioning evaluation},
  author={Sarto, Sara and Barraco, Manuele and Cornia, Marcella and Baraldi, Lorenzo and Cucchiara, Rita},
  booktitle={Proceedings of the IEEE/CVF conference on computer vision and pattern recognition},
  pages={6914--6924},
  year={2023}
}

@inproceedings{lee2024fleur,
  title={Fleur: An explainable reference-free evaluation metric for image captioning using a large multimodal model},
  author={Lee, Yebin and Park, Imseong and Kang, Myungjoo},
  booktitle={Proceedings of the 62nd Annual Meeting of the Association for Computational Linguistics (Volume 1: Long Papers)},
  pages={3732--3746},
  year={2024}
}

@article{Qwen3-VL,
      title={Qwen3-VL Technical Report}, 
      author={Shuai Bai and Yuxuan Cai and Ruizhe Chen and Keqin Chen and Xionghui Chen and Zesen Cheng and Lianghao Deng and Wei Ding and Chang Gao and Chunjiang Ge and Wenbin Ge and Zhifang Guo and Qidong Huang and Jie Huang and Fei Huang and Binyuan Hui and Shutong Jiang and Zhaohai Li and Mingsheng Li and Mei Li and Kaixin Li and Zicheng Lin and Junyang Lin and Xuejing Liu and Jiawei Liu and Chenglong Liu and Yang Liu and Dayiheng Liu and Shixuan Liu and Dunjie Lu and Ruilin Luo and Chenxu Lv and Rui Men and Lingchen Meng and Xuancheng Ren and Xingzhang Ren and Sibo Song and Yuchong Sun and Jun Tang and Jianhong Tu and Jianqiang Wan and Peng Wang and Pengfei Wang and Qiuyue Wang and Yuxuan Wang and Tianbao Xie and Yiheng Xu and Haiyang Xu and Jin Xu and Zhibo Yang and Mingkun Yang and Jianxin Yang and An Yang and Bowen Yu and Fei Zhang and Hang Zhang and Xi Zhang and Bo Zheng and Humen Zhong and Jingren Zhou and Fan Zhou and Jing Zhou and Yuanzhi Zhu and Ke Zhu},
	  journal={arXiv preprint arXiv:2511.21631},
      year={2025}
}

@misc{yao2025falconremotesensingvisionlanguage,
      title={Falcon: A Remote Sensing Vision-Language Foundation Model (Technical Report)}, 
      author={Kelu Yao and Nuo Xu and Rong Yang and Yingying Xu and Zhuoyan Gao and Titinunt Kitrungrotsakul and Yi Ren and Pu Zhang and Jin Wang and Ning Wei and Chao Li},
      year={2025},
      eprint={2503.11070},
      archivePrefix={arXiv},
      primaryClass={cs.CV},
      url={https://arxiv.org/abs/2503.11070}, 
}

@inproceedings{soni2025earthdial,
  title={Earthdial: Turning multi-sensory earth observations to interactive dialogues},
  author={Soni, Sagar and Dudhane, Akshay and Debary, Hiyam and Fiaz, Mustansar and Munir, Muhammad Akhtar and Danish, Muhammad Sohail and Fraccaro, Paolo and Watson, Campbell D and Klein, Levente J and Khan, Fahad Shahbaz and others},
  booktitle={Proceedings of the Computer Vision and Pattern Recognition Conference},
  pages={14303--14313},
  year={2025}
}

@article{chen2025integrating,
  title={Integrating Global and Local Information for Remote Sensing Image-Text Retrieval},
  author={Chen, Ziyun and Liu, Fan and Guan, Zhangqingyun and Zhou, Qian and Zhou, Xiaocong and Zhang, Chuanyi},
  journal={IEEE Geoscience and Remote Sensing Letters},
  year={2025},
  publisher={IEEE}
}

@inproceedings{yao2025remotesam,
  title={Remotesam: Towards segment anything for earth observation},
  author={Yao, Liang and Liu, Fan and Chen, Delong and Zhang, Chuanyi and Wang, Yijun and Chen, Ziyun and Xu, Wei and Di, Shimin and Zheng, Yuhui},
  booktitle={Proceedings of the 33rd ACM International Conference on Multimedia},
  pages={3027--3036},
  year={2025}
}

@inproceedings{kuckreja2024geochat,
  title={Geochat: Grounded large vision-language model for remote sensing},
  author={Kuckreja, Kartik and Danish, Muhammad Sohail and Naseer, Muzammal and Das, Abhijit and Khan, Salman and Khan, Fahad Shahbaz},
  booktitle={Proceedings of the IEEE/CVF conference on computer vision and pattern recognition},
  pages={27831--27840},
  year={2024}
}

@inproceedings{pang2025vhm,
  title={Vhm: Versatile and honest vision language model for remote sensing image analysis},
  author={Pang, Chao and Weng, Xingxing and Wu, Jiang and Li, Jiayu and Liu, Yi and Sun, Jiaxing and Li, Weijia and Wang, Shuai and Feng, Litong and Xia, Gui-Song and others},
  booktitle={Proceedings of the AAAI Conference on Artificial Intelligence},
  volume={39},
  number={6},
  pages={6381--6388},
  year={2025}
}

@article{luo2024skysensegpt,
  title={Skysensegpt: A fine-grained instruction tuning dataset and model for remote sensing vision-language understanding},
  author={Luo, Junwei and Pang, Zhen and Zhang, Yongjun and Wang, Tingzhu and Wang, Linlin and Dang, Bo and Lao, Jiangwei and Wang, Jian and Chen, Jingdong and Tan, Yihua and others},
  journal={arXiv preprint arXiv:2406.10100},
  year={2024}
}

@article{ou2025geopix,
  title={GeoPix: A multimodal large language model for pixel-level image understanding in remote sensing},
  author={Ou, Ruizhe and Hu, Yuan and Zhang, Fan and Chen, Jiaxin and Liu, Yu},
  journal={IEEE Geoscience and Remote Sensing Magazine},
  year={2025},
  publisher={IEEE}
}

@article{li2025describeearth,
  title={Describeearth: Describe anything for remote sensing images},
  author={Li, Kaiyu and Jiang, Zixuan and Cao, Xiangyong and Wang, Jiayu and Xiao, Yuchen and Meng, Deyu and Wang, Zhi},
  journal={arXiv preprint arXiv:2509.25654},
  year={2025}
}

@article{zhou2024geoground,
  title={GeoGround: A unified large vision-language model for remote sensing visual grounding},
  author={Zhou, Yue and Lan, Mengcheng and Li, Xiang and Feng, Litong and Ke, Yiping and Jiang, Xue and Li, Qingyun and Yang, Xue and Zhang, Wayne},
  journal={arXiv preprint arXiv:2411.11904},
  year={2024}
}

@article{Qwen2.5-VL,
  title={Qwen2.5-VL Technical Report},
  author={Bai, Shuai and Chen, Keqin and Liu, Xuejing and Wang, Jialin and Ge, Wenbin and Song, Sibo and Dang, Kai and Wang, Peng and Wang, Shijie and Tang, Jun and Zhong, Humen and Zhu, Yuanzhi and Yang, Mingkun and Li, Zhaohai and Wan, Jianqiang and Wang, Pengfei and Ding, Wei and Fu, Zheren and Xu, Yiheng and Ye, Jiabo and Zhang, Xi and Xie, Tianbao and Cheng, Zesen and Zhang, Hang and Yang, Zhibo and Xu, Haiyang and Lin, Junyang},
  journal={arXiv preprint arXiv:2502.13923},
  year={2025}
}

@article{liu2025text2earth,
  title={Text2Earth: Unlocking text-driven remote sensing image generation with a global-scale dataset and a foundation model},
  author={Liu, Chenyang and Chen, Keyan and Zhao, Rui and Zou, Zhengxia and Shi, Zhenwei},
  journal={IEEE Geoscience and Remote Sensing Magazine},
  year={2025},
  publisher={IEEE}
}

@misc{flux2024,
    author={Black Forest Labs},
    title={FLUX},
    year={2024},
    howpublished={\url{https://github.com/black-forest-labs/flux}},
}

@misc{flux-2-2025,
    author={Black Forest Labs},
    title={{FLUX.2: Frontier Visual Intelligence}},
    year={2025},
    howpublished={\url{https://bfl.ai/blog/flux-2}},
}

@article{team2025zimage,
  title={Z-Image: An Efficient Image Generation Foundation Model with Single-Stream Diffusion Transformer},
  author={Z-Image Team},
  journal={arXiv preprint arXiv:2511.22699},
  year={2025}
}

@article{yuan2022exploring,
  title={Exploring a fine-grained multiscale method for cross-modal remote sensing image retrieval},
  author={Yuan, Zhiqiang and Zhang, Wenkai and Fu, Kun and Li, Xuan and Deng, Chubo and Wang, Hongqi and Sun, Xian},
  journal={arXiv preprint arXiv:2204.09868},
  year={2022}
}

@inproceedings{shen2025chain,
  title={CHAIN-OF-TALKERS (COTALK): Fast Human Annotation of Dense Image Captions},
  author={Shen, Yijun and Chen, Delong and Liu, Fan and Wang, Xingyu and Zhang, Chuanyi and Yao, Liang and Zheng, Yuhui},
  booktitle={Proceedings of the 2025 Conference on Empirical Methods in Natural Language Processing},
  pages={4444--4464},
  year={2025}
}

@book{cover1999elements,
  title={Elements of information theory},
  author={Cover, Thomas M},
  year={1999},
  publisher={John Wiley \& Sons}
}

@article{kingma2013auto,
  title={Auto-encoding variational bayes},
  author={Kingma, Diederik P and Welling, Max},
  journal={arXiv preprint arXiv:1312.6114},
  year={2013}
}

@inproceedings{rezende2014stochastic,
  title={Stochastic backpropagation and approximate inference in deep generative models},
  author={Rezende, Danilo Jimenez and Mohamed, Shakir and Wierstra, Daan},
  booktitle={International conference on machine learning},
  pages={1278--1286},
  year={2014},
  organization={PMLR}
}

@article{blei2017variational,
  title={Variational inference: A review for statisticians},
  author={Blei, David M and Kucukelbir, Alp and McAuliffe, Jon D},
  journal={Journal of the American statistical Association},
  volume={112},
  number={518},
  pages={859--877},
  year={2017},
  publisher={Taylor \& Francis}
}

@article{banerjee2005clustering,
  title={Clustering on the Unit Hypersphere using von Mises-Fisher Distributions.},
  author={Banerjee, Arindam and Dhillon, Inderjit S and Ghosh, Joydeep and Sra, Suvrit and Ridgeway, Greg},
  journal={Journal of Machine Learning Research},
  volume={6},
  number={9},
  year={2005}
}

@inproceedings{wang2020understanding,
  title={Understanding contrastive representation learning through alignment and uniformity on the hypersphere},
  author={Wang, Tongzhou and Isola, Phillip},
  booktitle={International conference on machine learning},
  pages={9929--9939},
  year={2020},
  organization={PMLR}
}

@article{chen2024makes,
  title={What makes for good image captions?},
  author={Chen, Delong and Cahyawijaya, Samuel and Ishii, Etsuko and Chan, Ho Shu and Bang, Yejin and Fung, Pascale},
  journal={arXiv preprint arXiv:2405.00485},
  year={2024}
}

@inproceedings{urbanek2024picture,
  title={A picture is worth more than 77 text tokens: Evaluating clip-style models on dense captions},
  author={Urbanek, Jack and Bordes, Florian and Astolfi, Pietro and Williamson, Mary and Sharma, Vasu and Romero-Soriano, Adriana},
  booktitle={Proceedings of the IEEE/CVF Conference on Computer Vision and Pattern Recognition},
  pages={26700--26709},
  year={2024}
}

@inproceedings{muhtar2024lhrs,
  title={Lhrs-bot: Empowering remote sensing with vgi-enhanced large multimodal language model},
  author={Muhtar, Dilxat and Li, Zhenshi and Gu, Feng and Zhang, Xueliang and Xiao, Pengfeng},
  booktitle={European Conference on Computer Vision},
  pages={440--457},
  year={2024},
  organization={Springer}
}

@article{bazi2024rs,
  title={Rs-llava: A large vision-language model for joint captioning and question answering in remote sensing imagery},
  author={Bazi, Yakoub and Bashmal, Laila and Al Rahhal, Mohamad Mahmoud and Ricci, Riccardo and Melgani, Farid},
  journal={Remote Sensing},
  volume={16},
  number={9},
  pages={1477},
  year={2024},
  publisher={MDPI}
}

@article{zhang2024earthmarker,
  title={EarthMarker: A visual prompting multimodal large language model for remote sensing},
  author={Zhang, Wei and Cai, Miaoxin and Zhang, Tong and Zhuang, Yin and Li, Jun and Mao, Xuerui},
  journal={IEEE Transactions on Geoscience and Remote Sensing},
  volume={63},
  pages={1--19},
  year={2024},
  publisher={IEEE}
}

@article{lin2025rs,
  title={Rs-moe: A vision-language model with mixture of experts for remote sensing image captioning and visual question answering},
  author={Lin, Hui and Hong, Danfeng and Ge, Shuhang and Luo, Chuyao and Jiang, Kai and Jin, Hao and Wen, Congcong},
  journal={IEEE Transactions on Geoscience and Remote Sensing},
  year={2025},
  publisher={IEEE}
}

@article{wang2024ringmogpt,
  title={Ringmogpt: A unified remote sensing foundation model for vision, language, and grounded tasks},
  author={Wang, Peijin and Hu, Huiyang and Tong, Boyuan and Zhang, Ziqi and Yao, Fanglong and Feng, Yingchao and Zhu, Zining and Chang, Hao and Diao, Wenhui and Ye, Qixiang and others},
  journal={IEEE Transactions on Geoscience and Remote Sensing},
  volume={63},
  pages={1--20},
  year={2024},
  publisher={IEEE}
}

@article{liu2025boost,
  title={Boost uav-based object detection via scale-invariant feature disentanglement and adversarial learning},
  author={Liu, Fan and Yao, Liang and Zhang, Chuanyi and Wu, Ting and Zhang, Xinlei and Jiang, Xiruo and Zhou, Jun},
  journal={IEEE Transactions on Geoscience and Remote Sensing},
  year={2025},
  publisher={IEEE}
}

@article{yao2024domain,
  title={Domain-invariant progressive knowledge distillation for uav-based object detection},
  author={Yao, Liang and Liu, Fan and Zhang, Chuanyi and Ou, Zhiquan and Wu, Ting},
  journal={IEEE Geoscience and Remote Sensing Letters},
  volume={22},
  pages={1--5},
  year={2024},
  publisher={IEEE}
}

@inproceedings{yao2025uemm,
  title={UEMM-Air: Enable UAVs to Undertake More Multi-modal Tasks},
  author={Yao, Liang and Liu, Fan and Xu, Shengxiang and Zhang, Chuanyi and Di, Shimin and Ma, Xing and Jiang, Jianyu and Wang, Zequan and Zhou, Jun},
  booktitle={Proceedings of the 33rd ACM International Conference on Multimedia},
  pages={12792--12798},
  year={2025}
}

@article{liu2024remoteclip,
  title={Remoteclip: A vision language foundation model for remote sensing},
  author={Liu, Fan and Chen, Delong and Guan, Zhangqingyun and Zhou, Xiaocong and Zhu, Jiale and Ye, Qiaolin and Fu, Liyong and Zhou, Jun},
  journal={IEEE Transactions on Geoscience and Remote Sensing},
  volume={62},
  pages={1--16},
  year={2024},
  publisher={IEEE}
}

@inproceedings{radford2021learning,
  title={Learning transferable visual models from natural language supervision},
  author={Radford, Alec and Kim, Jong Wook and Hallacy, Chris and Ramesh, Aditya and Goh, Gabriel and Agarwal, Sandhini and Sastry, Girish and Askell, Amanda and Mishkin, Pamela and Clark, Jack and others},
  booktitle={International conference on machine learning},
  pages={8748--8763},
  year={2021},
  organization={PmLR}
}

@article{tschannen2025siglip,
  title={Siglip 2: Multilingual vision-language encoders with improved semantic understanding, localization, and dense features},
  author={Tschannen, Michael and Gritsenko, Alexey and Wang, Xiao and Naeem, Muhammad Ferjad and Alabdulmohsin, Ibrahim and Parthasarathy, Nikhil and Evans, Talfan and Beyer, Lucas and Xia, Ye and Mustafa, Basil and others},
  journal={arXiv preprint arXiv:2502.14786},
  year={2025}
}

@article{xie2025fg,
  title={FG-CLIP 2: A Bilingual Fine-grained Vision-Language Alignment Model},
  author={Xie, Chunyu and Wang, Bin and Kong, Fanjing and Li, Jincheng and Liang, Dawei and Ao, Ji and Leng, Dawei and Yin, Yuhui},
  journal={arXiv preprint arXiv:2510.10921},
  year={2025}
}

@article{oquab2023dinov2,
  title={Dinov2: Learning robust visual features without supervision},
  author={Oquab, Maxime and Darcet, Timoth{\'e}e and Moutakanni, Th{\'e}o and Vo, Huy and Szafraniec, Marc and Khalidov, Vasil and Fernandez, Pierre and Haziza, Daniel and Massa, Francisco and El-Nouby, Alaaeldin and others},
  journal={arXiv preprint arXiv:2304.07193},
  year={2023}
}

@article{simeoni2025dinov3,
  title={Dinov3},
  author={Sim{\'e}oni, Oriane and Vo, Huy V and Seitzer, Maximilian and Baldassarre, Federico and Oquab, Maxime and Jose, Cijo and Khalidov, Vasil and Szafraniec, Marc and Yi, Seungeun and Ramamonjisoa, Micha{\"e}l and others},
  journal={arXiv preprint arXiv:2508.10104},
  year={2025}
}

@article{fu2023dreamsim,
  title={Dreamsim: Learning new dimensions of human visual similarity using synthetic data},
  author={Fu, Stephanie and Tamir, Netanel and Sundaram, Shobhita and Chai, Lucy and Zhang, Richard and Dekel, Tali and Isola, Phillip},
  journal={arXiv preprint arXiv:2306.09344},
  year={2023}
}

@article{lu2017exploring,
  title={Exploring models and data for remote sensing image caption generation},
  author={Lu, Xiaoqiang and Wang, Binqiang and Zheng, Xiangtao and Li, Xuelong},
  journal={IEEE Transactions on Geoscience and Remote Sensing},
  volume={56},
  number={4},
  pages={2183--2195},
  year={2017},
  publisher={IEEE}
}

@article{liu2022remote,
  title={Remote-sensing image captioning based on multilayer aggregated transformer},
  author={Liu, Chenyang and Zhao, Rui and Shi, Zhenwei},
  journal={IEEE Geoscience and Remote Sensing Letters},
  volume={19},
  pages={1--5},
  year={2022},
  publisher={IEEE}
}

@article{gao2025multi,
  title={Multi-modal large models driven SAR image captioning: A benchmark dataset and baselines},
  author={Gao, Ziyi and Sun, Shuzhou and Cheng, Ming-Ming and Liu, Yongxiang and Liu, Li},
  journal={IEEE Journal of Selected Topics in Applied Earth Observations and Remote Sensing},
  year={2025},
  publisher={IEEE}
}

@inproceedings{zhang2025sc,
  title={Sc-captioner: Improving image captioning with self-correction by reinforcement learning},
  author={Zhang, Lin and Zeng, Xianfang and Li, Kangcong and Yu, Gang and Chen, Tao},
  booktitle={Proceedings of the IEEE/CVF International Conference on Computer Vision},
  pages={23145--23155},
  year={2025}
}

@inproceedings{peng2025patch,
  title={Patch Matters: Training-free Fine-grained Image Caption Enhancement via Local Perception},
  author={Peng, Ruotian and He, Haiying and Wei, Yake and Wen, Yandong and Hu, Di},
  booktitle={Proceedings of the IEEE/CVF Conference on Computer Vision and Pattern Recognition},
  pages={3963--3973},
  year={2025}
}

@inproceedings{zou2025remotetrimmer,
  title={Remotetrimmer: Adaptive structural pruning for remote sensing image classification},
  author={Zou, Guangwenjie and Yao, Liang and Liu, Fan and Zhang, Chuanyi and Li, Xin and Chen, Ning and Xu, Shengxiang and Zhou, Jun},
  booktitle={ICASSP 2025-2025 IEEE International Conference on Acoustics, Speech and Signal Processing (ICASSP)},
  pages={1--5},
  year={2025},
  organization={IEEE}
}

@inproceedings{jiang2026airnavigation,
  title={AirNavigation: Let UAV Navigation Tell Its Own Story},
  author={Jiang, Jianyu and Wang, Zequan and Yao, Liang and Xu, Shengxiang and Liu, Fan},
  booktitle={Proceedings of the AAAI Conference on Artificial Intelligence},
  volume={40},
  number={48},
  pages={41610--41612},
  year={2026}
}

@article{li2024language,
  title={Language-guided progressive attention for visual grounding in remote sensing images},
  author={Li, Ke and Wang, Di and Xu, Haojie and Zhong, Haodi and Wang, Cong},
  journal={IEEE Transactions on Geoscience and Remote Sensing},
  volume={62},
  pages={1--13},
  year={2024},
  publisher={IEEE}
}

@inproceedings{li2026rsvg,
  title={Rsvg-zeroov: Exploring a training-free framework for zero-shot open-vocabulary visual grounding in remote sensing images},
  author={Li, Ke and Wang, Di and Wang, Ting and Dong, Fuyu and Zhang, Yiming and Zhang, Luyao and Wang, Xiangyu and Li, Shaofeng and Wang, Quan},
  booktitle={Proceedings of the AAAI Conference on Artificial Intelligence},
  volume={40},
  number={8},
  pages={6288--6296},
  year={2026}
}

@article{yao2026remoteagent,
  title={RemoteAgent: Bridging Vague Human Intents and Earth Observation with RL-based Agentic MLLMs},
  author={Yao, Liang and Xu, Shengxiang and Liu, Fan and Zhang, Chuanyi and Yao, Bishun and Min, Rui and Li, Yongjun and Ouyang, Chaoqian and Di, Shimin and Zhang, Min-Ling},
  journal={arXiv preprint arXiv:2604.07765},
  year={2026}
}

\appendix

\end{document}